\documentclass{article}
\usepackage{iclr2017_workshop,times}
\usepackage{hyperref}
\usepackage{url}
\usepackage{subfig}
\usepackage{natbib}
\usepackage{graphicx}
\usepackage{amsmath}
\usepackage{microtype}
\usepackage{multicol}

\usepackage{booktabs}  

\title{Multiplicative LSTM for sequence modelling}

\author{Ben Krause, Iain Murray \& Steve Renals  \\
School of Informatics, University of Edinburgh\\
Edinburgh, Scotland, UK\\
\texttt{\{ben.krause,i.murray,s.renals\}@ed.ac.uk} \\
\And
Liang Lu \\
Toyota Technological Institute at Chicago \\
Chicago, Illinois, USA\\
\texttt{\{llu\}@ttic.edu} \\
}

\begin{document}

\maketitle

\begin{abstract}
We introduce multiplicative LSTM (mLSTM), a recurrent neural network architecture for sequence modelling that combines the long short-term memory (LSTM) and multiplicative recurrent neural network architectures. mLSTM is characterised by its ability to have different recurrent transition functions for each possible input, which we argue makes it more expressive for autoregressive density estimation. We demonstrate empirically that mLSTM outperforms standard LSTM and its deep variants for a range of character level language modelling tasks. In this version of the paper, we regularise mLSTM to achieve 1.27 bits/char on text8 and 1.24 bits/char on Hutter Prize. We also apply a purely byte-level mLSTM on the WikiText-2 dataset to achieve a character level entropy of 1.26 bits/char, corresponding to a word level perplexity of 88.8, which is comparable to word level LSTMs regularised in similar ways on the same task.
\end{abstract}
\section{Introduction}

Recurrent neural networks (RNNs) are powerful sequence density estimators that can use long contexts to make predictions.  They have achieved tremendous success in (conditional) sequence modelling tasks such as language modelling, machine translation and speech recognition.  Generative models of sequences can apply factorization via the product rule to perform density estimation of the sequence $x_{1:T} = \{x_1,\dots,x_T\}$,
\begin{equation}
P(x_1,\dots,x_T) = P(x_1) P(x_2|x_1)P(x_3|x_2,x_1)\cdots P(x_T|x_1\dots x_{T-1}).
\end{equation}
 RNNs can model sequences with the above factorization by using a hidden state to summarize past inputs. The hidden state vector $h_t$ is updated recursively using the previous hidden state vector $h_{t-1}$ and the current input $x_{t}$ as
\begin{equation}
h_t = \mathcal{F}(h_{t-1},x_t),
\end{equation}
where $\mathcal{F}$ is a differentiable function with learnable parameters. In a vanilla RNN, $\mathcal{F}$ multiplies its inputs by a matrix and squashes the result with a non-linear function such as a hyperbolic tangent ($\tanh$). The updated hidden state vector is then used to predict a probability distribution over the next sequence element, using function $\mathcal{G}$. In the case where $x_{1:T}$ consists of mutually exclusive discrete outcomes, $\mathcal{G}$ may apply a matrix multiplication followed by a softmax function:
\begin{equation}
P(x_{t+1}) = \mathcal{G}(h_t).
\end{equation}

Generative RNNs can evaluate log-likelihoods of sequences exactly, and are  differentiable with respect to these log-likelihoods. RNNs can be difficult to train due to the vanishing gradient problem \citep{Bengio-1994}, but advances such as the long short-term memory architecture (LSTM) \citep{Hochreiter-1997} have allowed RNNs to be successful. Despite their success, generative RNNs (as well as other conditional generative models) are known to have problems with recovering from mistakes \citep{Graves-2013}. Each time the recursive function of the RNN is applied and the hidden state is updated, the RNN must decide which information from the previous hidden state to store, due to its limited capacity. If the RNN's hidden representation remembers the wrong information and reaches a bad numerical state for predicting future sequence elements, for instance as a result of an unexpected input, it may take many time-steps to recover.

We argue that RNN architectures with hidden-to-hidden transition functions that are input-dependent are better suited to recover from surprising inputs. Our approach to generative RNNs combines LSTM units with multiplicative RNN (mRNN) factorized hidden weights, allowing flexible input-dependent transitions that are easier to control due to the gating units of LSTM \@. We compare this multiplicative LSTM hybrid architecture with other variants of LSTM on a range of character level language modelling tasks. Multiplicative LSTM is most appropriate when it can learn parameters specifically for each possible input at a given timestep. Therefore, its main application is to sequences of discrete mutually exclusive elements, such as language modelling and related problems.

\subsection{Input-dependent transition functions}

RNNs learn a mapping from previous hidden state $h_{t-1}$ and input $x_t$ to hidden state $h_t$. Let $\hat{h}_t$ denote the input to the next hidden state before any non-linear operation:
\begin{equation}
\hat{h}(t) = W_{hh}h_{t-1} + W_{hx}x_{t}  ,
\end{equation}
where  $W_{hh}$ is the hidden-to-hidden weight matrix, and $W_{hx}$ is the input-to-hidden weight matrix. For problems such as language modelling, $x_t$ is a one-hot vector, meaning that the output of $W_{hx}x_{t}$ is a column in $W_{hx}$, corresponding to the unit element in $x_{t}$.

\begin{figure}[tb]
\includegraphics[width=1.0\textwidth]{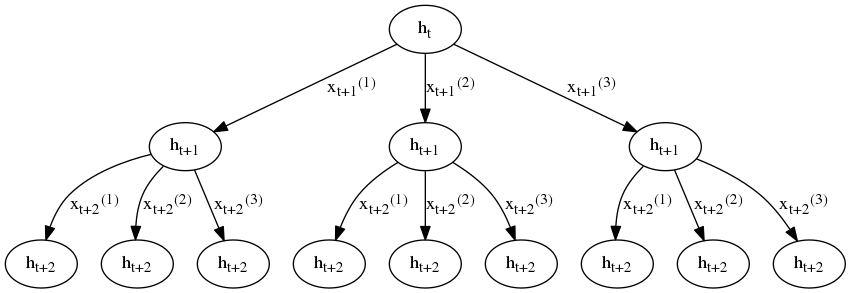}
\caption{Diagram of hidden states of a generative RNN as a tree, where $x_{t}^{(n)}$ denotes which of $N$ possible inputs is encountered at timestep $t$. Given $h_t$, the starting node of the tree, there will be a different possible $h_{t+1}$ for every $x_{t+1}^{(n)}$.  Similarly, for every $h_{t+1}$ that can be reached from  $h_t$, there is a different possible $h_{t+2}$ for each $x_{t+2}^{(n)}$, and so on.}
\label{fig:tree}
\end{figure}

The possible future hidden states in an RNN can be viewed as a tree structure, as shown in Figure \ref{fig:tree}. For an alphabet of $N$ inputs and a fixed $h_{t-1}$, there will be $N$ possible transition functions between $h_{t-1}$ and $\hat{h}_t$. The relative magnitude of $W_{hh}h_{t-1}$ to $W_{hx}x_{t}$ will need to be large for the RNN to be able to use long range dependencies, and the resulting possible hidden state vectors will therefore be highly correlated across the possible inputs, limiting the width of the tree and making it harder for the RNN to form distinct hidden representations for different sequences of inputs. However, if the RNN has flexible input-dependent transition functions,  the tree will be able to grow wider more quickly, giving the RNN the flexibility to represent more probability distributions.

In a vanilla RNN, it is difficult to allow inputs to greatly affect the hidden state vector without erasing information from the past hidden state. However,  an RNN with a transition function mapping $\hat{h}_t \leftarrow h_{t-1}$ dependent on the input would allow the relative values of $h_{t}$ to vary with each possible input $x_{t}$, without overwriting the contribution from the previous hidden state, allowing for more long term information to be stored. This ability to adjust to new inputs quickly while limiting the overwriting of information should make an RNN more robust to mistakes when it encounters surprising inputs, as the hidden vector is less likely to get trapped in a bad numerical state for making future predictions.  

\subsection{Multiplicative RNN}
The multiplicative RNN (mRNN) \citep{Sutskever-2011} is an architecture designed specifically to allow flexible input-dependent transitions. Its formulation was inspired by the tensor RNN, an RNN architecture that allows for a different transition matrix for each possible input. The tensor RNN features a 3-way tensor $W_{hh}^{1:N}$, which contains a separately learned transition matrix $W_{hh}$ for each input dimension. The 3-way tensor can be stored as an array of matrices 
\begin{equation}
W_{hh}^{(1:N)} = \{ W_{hh}^{(1)},...,W_{hh}^{(N)} \},
\end{equation}
where superscript is used to denote the index in the array, and $N$ is the dimensionality of $x_t$. The specific hidden-to-hidden weight matrix  $W_{hh}^{(x_t)}$ used for a given input $x_t$ is then
\begin{equation}
 W_{hh}^{(x_t)} = \sum_{n=1}^{N} W_{hh}^{(n)} x_t^{(n)}.
\end{equation}
For language modelling problems, only one unit of $x_t$ will be on, and  $W_{hh}^{(x_t)}$ will be the matrix in $W_{hh}^{(1:N)}$ corresponding to that unit. Hidden-to-hidden propagation in the tensor RNN is then given by
\begin{equation}
\hat{h}(t) =  W_{hh}^{(x_t)} h_{t-1} + W_{hx}x_{t}.
\end{equation}

The large number of parameters in the tensor RNN make it impractical for most problems. mRNNs can be thought of as a shared-parameter approximation to the tensor RNN that use a factorized hidden-to-hidden transition matrix in place of the normal RNN hidden-to-hidden matrix $W_{hh}$, with an input-dependent intermediate diagonal matrix $\mathrm{diag}(W_{mx}x_t)$. The input-dependent hidden-to-hidden weight matrix, $W_{hh}^{(x_t)}$ is then
\begin{equation}
W_{hh}^{(x_t)} = W_{hm} \mathrm{diag}(W_{mx}x_t) W_{mh}.
\end{equation}

An mRNN is thus equivalent to a tensor RNN using the above form for $W_{hh}^{(x_t)}$. For readability, an mRNN can also be described using intermediate state $m_t$ as follows:
\begin{align}
m_t &= (W_{mx} x_t) \odot (W_{mh}h_{t-1}) \\
\hat{h}_t &= W_{hm}m_t + W_{hx}x_{t}.
\end{align}

mRNNs have improved on vanilla RNNs at character level language modelling tasks \citep{Sutskever-2011,mikolov2012c}, but have fallen short of the more popular LSTM architecture, for instance as shown with LSTM baselines from \citep{cooijmans2017}. The standard RNN units in an mRNN do not provide an easy way for information to bypass its complex transitions, resulting in the potential for difficulty in retaining long term information.

\subsection{Long short-term memory}
LSTM is a commonly used RNN architecture that uses a series of multiplicative gates to control how information flows in and out of internal states of the network \citep{Hochreiter-1997}. There are several slightly different variants of LSTM, and we present the variant used in our experiments. 

The LSTM hidden state receives inputs from the input layer $x_t$ and the previous hidden state $h_{t-1}$:
\begin{equation}
\hat{h}_t= W_{hx}x_t + W_{hh}h_{t-1}.
\end{equation}

 The LSTM network also has 3 gating units -- input gate $i$, output gate $o$, and forget gate $f$ -- that have both recurrent and feed-forward connections:
\begin{align}
i_t &= \sigma(W_{ix}x_t + W_{ih}h_{t-1}) \\
o_t &= \sigma(W_{ox}x_t + W_{oh}h_{t-1}) \\
f_t &= \sigma(W_{fx}x_t + W_{fh}h_{t-1}),
\end{align}
where $\sigma$ is the logistic sigmoid function. The input gate controls how much of the input to each hidden unit is written to the internal state vector $c_t$, and the forget gate determines how much of the previous internal state $c_{t-1}$ is preserved. This combination of write and forget gates allows the network to control what information should be stored and overwritten across each time-step. The internal state is updated by
\begin{equation}
c_{t} = f_t \odot c_{t-1} + i_t \odot tanh(\hat{h}_t).
\end{equation}

The output gate controls how much of each unit's activation is preserved. It allows the LSTM cell to keep information that is not relevant to the current output, but may be relevant later. The final output of the hidden state is given by
\begin{equation}
  h_t = \tanh(c_t)\odot o_t.
\end{equation}
LSTM's ability to control how information is stored in each unit has proven generally useful.

\subsection{Comparing LSTM with mRNN}

The LSTM and mRNN architectures both feature multiplicative units, but these units serve different purposes. LSTM's gates are designed to control the flow of information through the network, whereas mRNN's gates are designed to allow transition functions to vary across inputs. LSTM gates receive input from both the input units and hidden units, allowing multiplicative interactions between hidden units, but also potentially limiting the extent of input-hidden multiplicative interaction. LSTM gates are also squashed with a sigmoid, forcing them to take values between 0 and 1, which makes them easier to control, but less expressive than mRNN's linear gates. For language modelling problems, mRNN's linear gates do not need to be controlled by the network because they are explicitly learned for each input. They are also placed in between a product of 2 dense matrices, giving more flexibility to the possible values of the final product of matrices.

\section{Multiplicative LSTM}
Since the LSTM and mRNN architectures are complimentary, we propose the multiplicative LSTM (mLSTM), a hybrid architecture that combines the factorized hidden-to-hidden transition of mRNNs with the gating framework from LSTMs. The mRNN and LSTM architectures can be combined by adding connections from the mRNN's intermediate state $m_t$ (which is redefined below for convenience) to each gating units in the LSTM, resulting in the following system:
\begin{align}
m_t &= (W_{mx} x_t) \odot (W_{mh}h_{t-1}) \\
\hat{h}_t &= W_{hx}x_t + W_{hm}m_{t} \\
i_t &= \sigma(W_{ix}x_t + W_{im}m_{t}) \\
o_t &= \sigma(W_{ox}x_t + W_{om}m_{t}) \\
f_t &= \sigma(W_{fx}x_t + W_{fm}m_{t}).
\end{align}

We set the dimensionality of $m_t$ and $h_t$ equal for all our experiments. We also chose to share $m_t$ across all LSTM unit types, resulting in a model with 1.25 times the number of recurrent weights as LSTM for the same number of hidden units.

The goal of this architecture is to combine the flexible input-dependent transitions of mRNNs with the long time lag and information control of LSTMs. The gated units of LSTMs could make it easier to control (or bypass) the complex transitions in that result from the factorized hidden weight matrix. The additional sigmoid input and forget gates featured in LSTM units allow even more flexible input-dependent transition functions than in regular mRNNs. 

\section{Related approaches}
Many recently proposed RNN architectures use recurrent depth, which is depth between recurrent steps. Recurrent depth allows more non-linearity in the combination of inputs and previous hidden states from every time step, which in turn allows for more flexible input-dependent transitions. Recurrent depth has been found to perform better than other kinds of non-recurrent depth for sequence modelling \citep{zhang2016}. Recurrent highway networks (RHNs) \citep{zilly2017} use a more sophisticated recurrent depth that carefully controls propagation through layers using gating units. The gating units also allow for a greater deal of multiplicative interaction between the inputs and hidden units. While adding recurrent depth could improve our model, we believe that maximizing the input-dependent flexibility of the transition function is more important for expressive sequence modelling. Recurrent depth can do this through non-linear layers combining hidden and input contributions, but our method can do this independently of non-linear depth.

 Another approach, multiplicative integration RNNs (MI-RNNs) \citep{wu2016}, use Hadamard products instead of addition when combining contributions from input and hidden units. When applying this to LSTM, this architecture achieves impressive sequence modelling results. The main difference between multiplicative integration LSTM and mLSTM is that mLSTM applies the Hadamard product between the multiplication of two matrices. In the case of LSTM, this allows for the potential for greater expressiveness, without significantly increasing the size of the model.

\section{Experiments}

\subsection{System Setup}

Our experiments measure the performance of mLSTM for character-level language modelling tasks of varying complexity\footnote{Code to replicate our experiments on the Hutter Prize dataset is available at \url{https://github.com/benkrause/mLSTM}.}. Our initial experiments, which appeared in previous versions of this work, were mainly designed to compare the convergence and final performance of mLSTM vs LSTM and its deep variants. Our follow up experiments explored training and regularisation of mLSTM in more detail, with goal of comparing more directly with the most competitive architectures in the literature. 

Our initial and follow up experiments used slightly different set ups; initial experiments used a variant of RMSprop, \citep{tieleman2012}, with normalized updates in place of a learning rate. All unnormalized update directions $v_*$, computed by RMSprop, were normalized to have length $\ell$, where $\ell$ was decayed exponentially over training:
\begin{equation}
v \leftarrow \frac{\ell}{\sqrt{v_*^T v_*}} v_*  .
\end{equation}

This update rule is similar to applying gradient norm clipping \citep{pascanu2013}, with a very high learning rate balanced out by a very low gradient norm threshold. The initial experiments also used a slightly non-standard version of LSTM (and mLSTM) with the output gate inside of the final tanh of the LSTM cell. This gave us slightly better results in preliminary experiments with very small models, but likely does not make much difference. We use LSTM (RMSprop) and mLSTM (RMSprop) in tables to distinguish results obtained by these initial set of experiments.

For our follow up experiments, we use more standard methodology to be more comparable to the literature. We used ADAM \citep{kingma2014}, always starting with an initial learning rate of $0.001$ and decaying this linearly to a minimum learning rate (which was always in the range $0.00005$ to $0.0001$). The mLSTMs used the standard LSTM cell with the output gate outside the tanh. These mLSTMs also used scaled orthogonal initialisations \citep{saxe2013} for the hidden weights, an initial forget gate bias of 3, and truncated backpropogation lengths from 200 to 250.

We compared mLSTM to previously reported regular LSTM, stacked LSTM, and RNN character-level language models. We run detailed experiments on the text8 and Hutter Prize datasets \citep{Hutter2012} to test medium scale character-level language modelling. We test our best model from these experiments on the WikiText-2 dataset \citep{Merity2016} to measure performance on smaller scale character level language modelling, and to compare with word level models. Previous versions of the paper also report a character level result on Penn Treebank dataset \citep{marcus1993} of 1.35 bits/char with an unregularised mLSTM, however we do not include this experiment in this version as we have no results with our updated training and regularisation methodology.

\subsection{Hutter Prize dataset}
We performed experiments using the Hutter Prize dataset, originally used for the Hutter Prize compression benchmark \citep{Hutter2012}. This dataset consists mostly of English language text and mark-up language text, but also contains text in other languages, including non-Latin languages. The dataset is modelled using a UTF-8 encoding, and contains 205 unique bytes. 

\begin{figure}
  \begin{minipage}[b]{0.5\textwidth}
    \includegraphics[width=\textwidth]{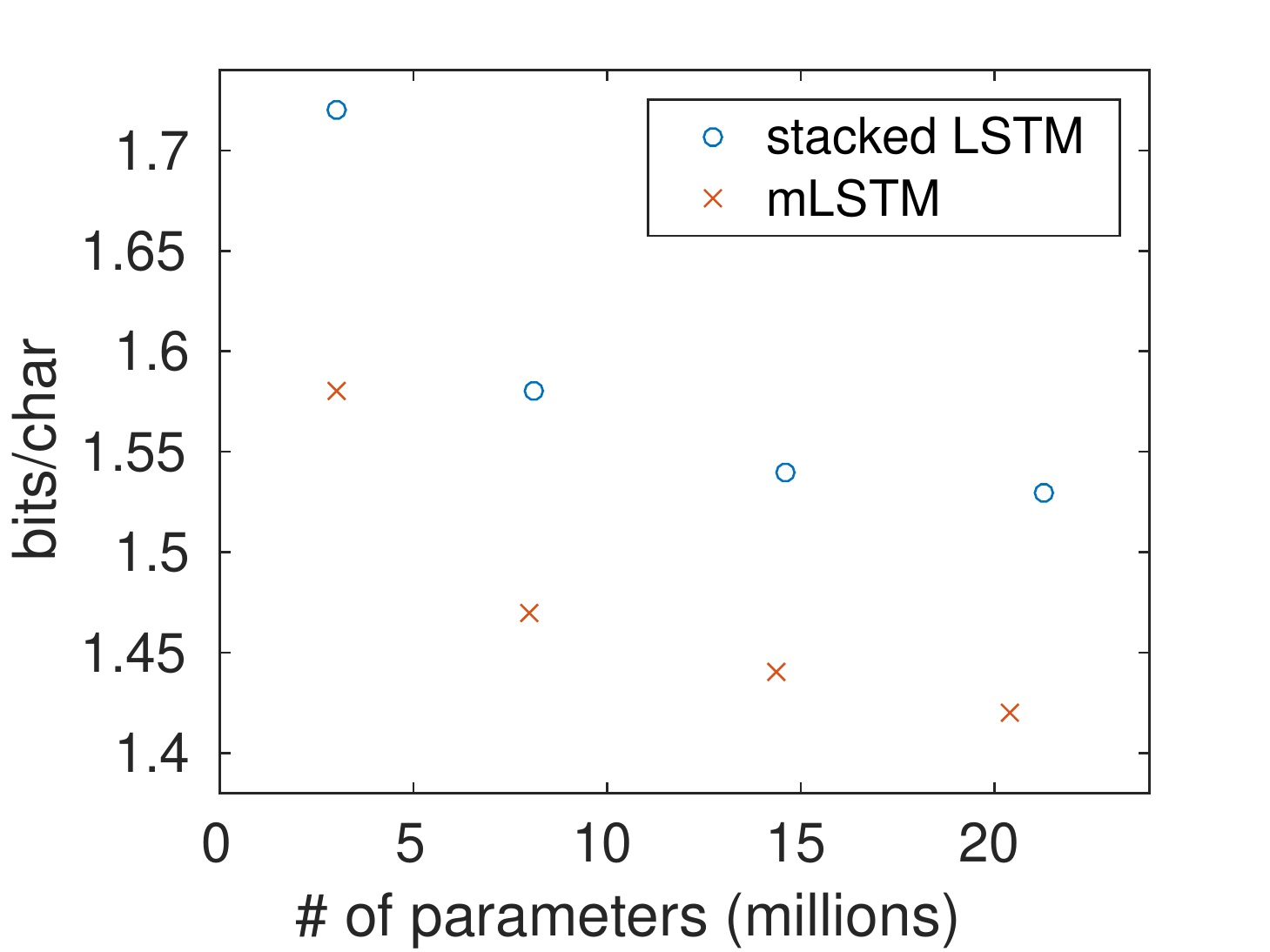}
    \caption{Hutter Prize validation performance \\ in bits/char plotted against number of network \\ parameters for mLSTM and stacked LSTM.}
    \label{fig:hutter-res}
  \end{minipage}
  \begin{minipage}[b]{0.5\textwidth}
    \includegraphics[width=\textwidth]{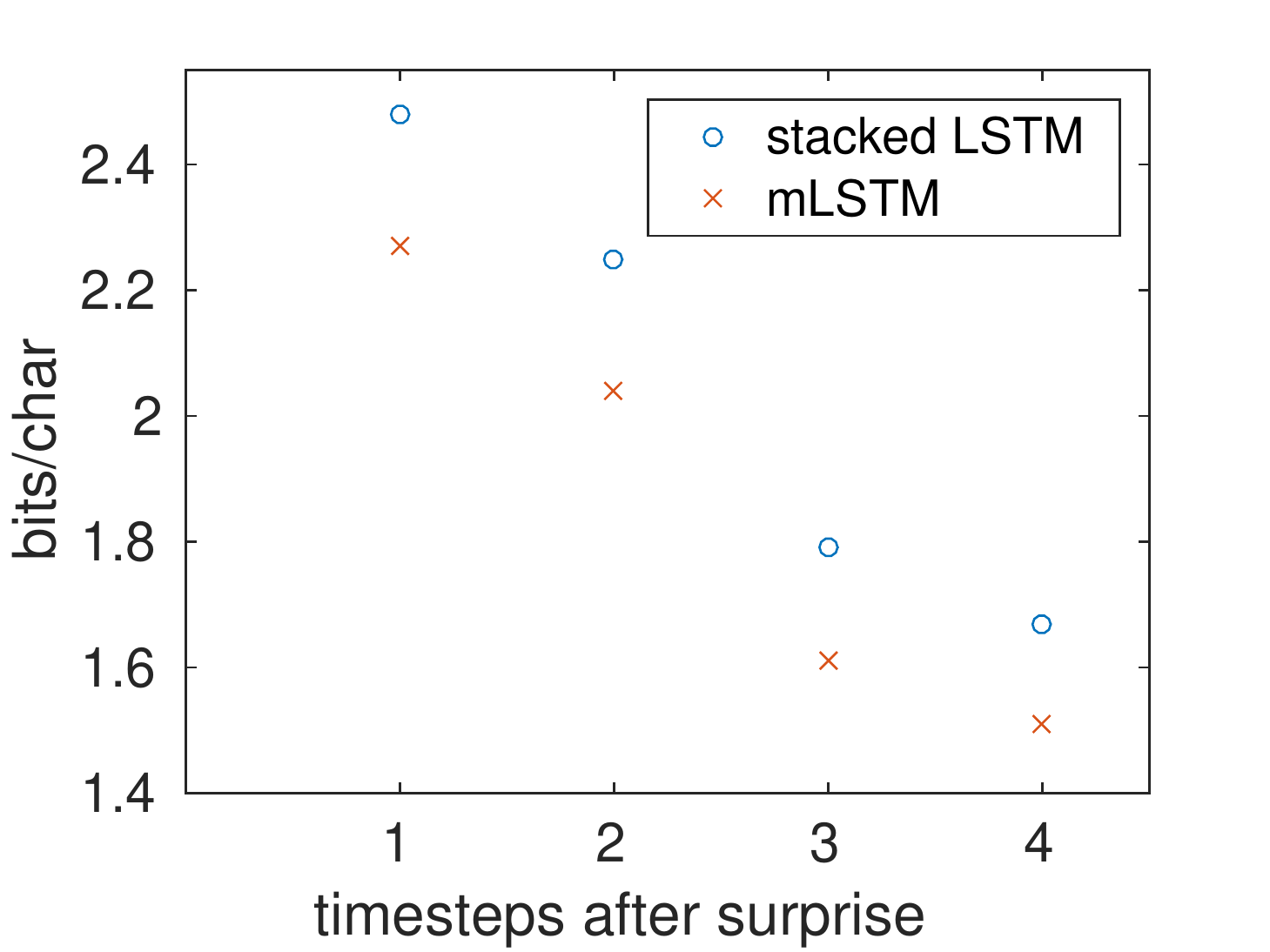}
    \caption{Cross entropy loss for mLSTM and stacked LSTM immediately proceeding a \\ surprising input}
    \label{fig:surprise-res}
  \end{minipage}
\end{figure}

In our initial experiments, we compared mLSTMs and 2-layer stacked LSTMs for varying network sizes, ranging from about 3--20 million parameters. These results all used RMS prop with normalized updates, stopping after 4 epochs on the first 95 million characters, with test performance measured on the last 5 million bytes. Hyperparameters for each mLSTM and stacked LSTM were kept constant across all sizes. The results, shown in Figure~\ref{fig:hutter-res}, show that mLSTM gives an improvement across all network sizes. 

We hypothesized that mLSTM's superior performance over stacked LSTM was in part due to its ability to recover from surprising inputs. To test this we looked at each network's performance after viewing surprising inputs that occurred naturally in the test set by creating a set of the 10\% characters with the largest average loss taken by mLSTM and stacked LSTM. Both networks perform roughly equally on this set of surprising characters, with mLSTM and stacked LSTM taking losses of 6.27 bits/character and 6.29 bits/character respectively. However, stacked LSTM tended to take much larger losses than mLSTM in the timesteps immediately following surprising inputs. One to four time-steps after a surprising input occurred, mLSTM and stacked LSTM took average losses of (2.26, 2.04, 1.61, 1.51) and (2.48, 2.25, 1.79, 1.67) bits per character respectively, as shown in Figure \ref{fig:surprise-res}. mLSTM's overall advantage over stacked LSTM was 1.42 bits/char to 1.53 bits/char; mLSTM's advantage over stacked LSTM was greater after a surprising input than it is in general. 

We also explore more standard training methodology and regularisation methods on this dataset. These experiments all used ADAM, and the standard 90-5-5 training validation test split on this dataset. We firstly consider a standard unregularised mLSTM trained with this methodology. We then experiment with an mLSTM with a linear embedding layer and weight normalization \citep{salimans2016} on recurrent weights (mLSTM +emb +WN), which is similar to the mLSTM architecture used in~\citep{radford2017}, which was built off our initial work. We also consider regularisation of the later model with variational dropout \citep{gal2016} (mLSTM +emb +WN +VD). Variational dropout is a form of dropout \citep{srivastava2014} where the dropout mask is shared across a sequence.

The standard unregularised LSTM used 1900 hidden units and 20 million parameters. The weight normalized mLSTM used 1900 hidden units, and a linear embedding layer of 400, giving it 22 million parameters. The large embedding layer was used because it was found to work well with dropout. Since this embedding layer is linear, it could potentially be removed during test time by multiplying its incoming and outgoing weight matrices to reduce the number of parameters (however we report parameter numbers with the embedding layer). For the regularised weight normalized mLSTM, we apply a variational dropout of 0.2 to the hidden state and to the embedding layer (dropout masks for both the hidden state and embedding layer were shared across a sequence). We also consider a larger version of the weight normalized mLSTM with 2800 hidden units and 46 million parameters. We increased the dropout in the embedding layer to 0.5 on this model. All results without variational dropout used early stopping on the validation error to reduce overfitting. The results for these experiments are given in table \ref{tab:wiki-res}.

\begin{table}[h]
\begin{center} 
\begin{tabular}{  l  l  l  l  l  l  l } \toprule 
architecture & \# of parameters & test set error \\ 
\midrule 
stacked LSTM (7-layer) \citep{Graves-2013} &  21M & 1.67 \\ 
stacked LSTM (7-layer) + dynamic eval \citep{Graves-2013} &  21M & 1.33 \\ 
MI-LSTM \citep{wu2016} &  17M & 1.44 \\  
recurrent memory array structures \citep{rocki2016} & & 1.40 \\
feedback LSTM + zoneout  \citep{rocki2016b} & & 1.37 \\
hyperLSTM  \citep{Ha2017} & 27 M & 1.34\\   
hierarchical multiscale LSTM \citep{chung2017} & & 1.32 \\  
bytenet decoder \citep{Kalchbrenner2016} & & 1.31\\  
LSTM (4 layer) + VD + BB tuning \citep{melis2017}& 46M & 1.30\\
RHN (rec depth 7) + VD \citep{zilly2017} & 46M & 1.27\\
Fast-slow LSTM (rec depth 4)  + zoneout \citep{mujika2017} & 47M & 1.25\\
\midrule
\textbf{unregularised mLSTM (RMS prop, 4 epoch)} & \textbf{20M} & \textbf{1.42} \\
\textbf{unregularised mLSTM} & \textbf{20M} & \textbf{1.40} \\
\textbf{mLSTM +emb +WN}  & \textbf{22M} & \textbf{1.44} \\
\textbf{mLSTM +emb +WN +VD} & \textbf{22M} & \textbf{1.28} \\
\textbf{large mLSTM +emb +WN +VD}  & \textbf{46M} & \textbf{1.24} \\
\bottomrule
\end{tabular} 
\end{center}
\caption{Hutter Prize dataset test error in bits/char.}
\label{tab:wiki-res}
\end{table}
Interestingly, adding weight normalization and an embedding layer hurt performance in the absence of regularisation. However, when combined with variational dropout, this model outperformed all previous static single model neural network results on Hutter Prize. We did not explore variational dropout applied to mLSTM without weight normalization. Earlier versions of this work also considered dynamic evaluation of mLSTMs on this task, however this is now in a separate paper focused on dynamic evaluation \citep{krause2017}.
 
We also tested an MI-LSTM, mLSTM's nearest neighbor, with a slightly larger size (22M parameters) and a very similar hyperparameter configuration and initialisation scheme~\footnote{The only difference in settings was the scale for the orthogonally initialised hidden weights; mLSTM used 0.7 and MI-LSTM used 0.5. We believed this was justified because mLSTM uses a product of two matrices, resulting in a spectral radius of 0.49 for this product. Additionally, reducing the scale to 0.5 improved MI-LSTM's initial convergence rate. Downscaling the orthogonal initialisations was necessary in general because an initial forget gate bias of 3 was used.} (compared with unregularised mLSTM with no WN). MI-LSTM achieved a relatively poor test set performance of 1.53 bits/char, as compared with 1.40 bits/char for mLSTM under the same settings. The MI-LSTM also converged more slowly, although eventually did require early stopping like the mLSTM. While this particular experiment cannot conclusively prove anything about the relative utility of mLSTM vs. MI-LSTM on this task, it does show that the two architectures are sufficiently different to obtain very different results under the same hyper-parameter settings.
\subsection{Text8 dataset}
Text8 contains 100 million characters of English text taken from Wikipedia in 2006, consisting of just the 26 characters of the English alphabet plus spaces. This dataset can be found at \url{http://mattmahoney.net/dc/textdata}. This corpus has been widely used to benchmark RNN character level language models, with the first 90 million characters used for training, the next 5 million used for validation, and the final 5 million used for testing.  The results of these experiments are shown in Table~\ref{tab:text8-res}.

The first set of experiments we performed were designed to be comparable to those of \citet{zhang2016}, who benchmarked several deep LSTMs against shallow LSTMs on this dataset. The shallow LSTM had a hidden state dimensionality of 512, and the deep versions had reduced dimensionality to give them roughly the same number of parameters. Our experiment used an mLSTM with a hidden dimensionality of 450, giving it slightly fewer parameters than the past work, and our own LSTM baseline with hidden dimensionality 512. mLSTM showed an improvement over our baseline and the previously reported best deep LSTM variant. 

We also ran experiments to compare a large mLSTM with other reported experiments. We trained an mLSTM with hidden dimensionality of 1900 on the text8 dataset. Unregularised mLSTM was able to fit the training data well and  achieved a competitive performance; however it was outperformed by other architectures that are less prone to over-fitting. 

We later considered our best training setup from the Hutter Prize dataset, reusing the exact same architecture and hyper-parameters from this task, with the only difference being the number of input characters (27 for text8), which reduces the number of parameters to around 45 million. This well regularised mLSTM was able to achieve a much stronger performance on text8, tying RHNs with a recurrent depth of 10 for the best result on this dataset.

\begin{table}[h]
\begin{center} 
\begin{tabular}{  l  l  l  l  l  l  l } \toprule 
architecture & test set error \\ 
\midrule 
mRNN \citep{mikolov2012c} & 1.54 \\
MI-LSTM \citep{wu2016} & 1.44 \\
LSTM \citep{cooijmans2017} & 1.43 \\
batch normalised LSTM \citep{cooijmans2017} & 1.36 \\
layer-norm hierarchical multiscale LSTM \citep{chung2017}  & 1.29 \\
Recurrent highway networks, rec. depth 10 +VD \citep{zilly2017} & 1.27 \\
\midrule 
small LSTM \citep{zhang2016} & 1.65 \\
small deep LSTM (best) \citep{zhang2016}  & 1.63 \\
\textbf{small LSTM (RMSprop)} & \textbf{1.64} \\
\textbf{small mLSTM (RMSprop)}  & \textbf{1.59}  \\
\midrule 
\textbf{unregularised mLSTM (RMSprop)} & \textbf{1.40} \\
\textbf{large mLSTM +emb +WN +VD} & \textbf{1.27} \\
\bottomrule
\end{tabular} 
\end{center}
\caption{Text8 dataset test set error in bits/char. Architectures labelled with small used a highly restrictive hidden dimensionality (512 for LSTM, 450 for mLSTM).}
\label{tab:text8-res}
\end{table}

\subsection{WikiText-2}
The WikiText-2 dataset has been a common benchmark for very recent advances in word-level language modelling. This dataset contains 2 million training tokens and a vocab size of 33k. Documents are given in non-shuffled order, causing the data to contain more long-range dependencies. We use this dataset to benchmark how our advances in character-level language modelling stack up against word level language models. Character language models generally perform worse than word-level language models on standard English text benchmarks. One reason for this is word level language models know the test set vocabulary in advance, whereas character level models model a distribution over all possible words, including out of vocabulary words, making the task inherently more difficult from character level view. Furthermore, very rare words, which character level models are more equipped to handle than word level models, are mapped to an unknown token. From the perspective of training, character level language models must model longer range dependencies, and must learn a more complex non-linear fit to capture joint dependencies between characters. Character level models do have an inherent advantage of being able to capture subword language information, motivating their use on traditionally word-level tasks.

Character level language models can be compared with word level language models by converting bits per character to perplexity. In this case, we model the data at the UTF-8 byte level. The bits per word can be computed as
\begin{equation}
bits/word = bits/symbol \times \frac{symbols/file}{words/file}
\end{equation}
where in this case, symbols are UTF-8 bytes. 2 raised to the power of the number of bits/word is then the perplexity. The WikiText-2 test set is 245,569 words long, and 1,256,449 bytes long, so each word is on average 5.1165 UTF-8 bytes long. A character level model can also assign word level probabilities directly by taking the product of the probabilities of the characters in a word, including the probability of the character ending the word (either a space or a newline). A byte level model is likely at a slight disadvantage compared with word-level because it must predict some information that gets removed during tokenization (such as spaces vs. newlines), but the perplexity given by the conversion above could atleast be seen as an upper bound of the word level perplexity such a model could achieve predicting byte by byte. This is because the entropy of the file after tokenization (which word level models measure) will always be less than or equal to the entropy of the file before tokenization (which byte level models measure).

We trained the mLSTM configuration from the Hutter Prize dataset, using an embedding layer, weight normalization, and a variational dropout of 0.5 in both the hidden and embedding layer, to model WikiText-2 at the byte level. This model contained 46 million parameters, which is larger than most word level models that use tied input and output embeddings \citep{press2017,inan2017} to share parameters, but similar in size to untied word level models on this dataset. The results are given in table~\ref{tab:wikitext}.
\begin{table}[h]
\begin{center} 
\begin{tabular}{  l  l  l  l  l  l  l } \toprule 
architecture &  valid & test  \\ 
\midrule 
LSTM   \citep{grave2017} & 104.2 & 99.3 \\
LSTM + VD (untied)\citep{inan2017} & 98.8 & 93.1 \\
LSTM + VD (tied)\citep{inan2017} & 91.5 & 87.0 \\
Pointer Sentinel LSTM \citep{Merity2016} & 84.8 & 80.8 \\
LSTM (tied) + VD + BB tuning \citep{melis2017} & 69.1 & 65.9 \\
LSTM + neural cache \citep{grave2017} & 72.1 & 68.9 \\
LSTM + dynamic eval \citep{krause2017} &  63.7 & 59.8\\
AWD-LSTM (tied) \citep{merity2017} & 68.6 & 65.8 \\
AWD-LSTM (tied)+ neural cache \citep{merity2017} & 53.8 & 52.0 \\
AWD-LSTM (tied) + dynamic eval \citep{krause2017} & 46.4 & 44.3 \\
\midrule 
\textbf{byte mLSTM +emb +WN +VD} & \textbf{92.8} & \textbf{88.8} \\
\bottomrule
\end{tabular} 
\end{center}
\caption{WikiText-2 perplexity errors}
\label{tab:wikitext}
\end{table}

Byte mLSTM achieves a byte-level test set cross entropy of $1.2649$ bits/char, corresponding to a perplexity of $88.8$. Despite all the disadvantages faced by character level models, byte level mLSTM achieves similar word level perplexity to previous word-level LSTM baselines that also use variational dropout for regularisation. Byte mLSTM does not perform as well as word-level models that use adaptive add-on methods or very recent advances in regularisation/hyper-parameter tuning, however it could likely benefit from these advances as well.

\section{Discussion}
This work combined the mRNN's factorized hidden weights with the LSTM's hidden units for generative modelling of discrete multinomial sequences. This mLSTM architecture was motivated by its ability to have both controlled and flexible input-dependent transitions, to allow for fast changes to the distributed hidden representation without erasing information. In a series of character-level language modelling experiments, mLSTM showed improvements over LSTM and its deep variants. mLSTM regularised with variational dropout performed favorably compared with baselines in the literature, outperforming all previous neural models on Hutter Prize and tying the best previous result on text8. Byte-level mLSTM was also able to perform competitively with word-level language models on WikiText-2.

Unlike many previous approaches that have achieved success at character level language modelling, mLSTM does not use non-linear recurrent depth. All mLSTMs considered in this work only had 2 linear recurrent transition matrices, whereas comparable works such as recurrent highway networks use a recurrent depth of up to 10 to achieve best results. This makes mLSTM more easily parallelizable than these approaches. Additionally, our work suggests that a large depth is not necessary to achieve competitive results on character level language modelling. We hypothesize that mLSTM's ability to have very different transition functions for each possible input is what makes it successful at this task. While recurrent depth can accomplish this too, mLSTM can achieve this more efficiently.

While these results are promising, it remains to be seen how mLSTM performs at word-level language modelling and other discrete multinomial generative modelling tasks, and whether mLSTM can be formulated to apply more broadly to tasks with continuous or non-sparse input units. We also hope this work will motivate further exploration in generative RNN architectures with flexible input-dependent transition functions.

\bibliography{iclr2017_conference}
\bibliographystyle{iclr2017_conference}

\end{document}